\documentclass[acmtog]{acmart}

\renewcommand\footnotetextcopyrightpermission[1]{} 
\AtBeginDocument{%
  \providecommand\BibTeX{{%
    \normalfont B\kern-0.5em{\scshape i\kern-0.25em b}\kern-0.8em\TeX}}}

\setcopyright{none}

\acmSubmissionID{papers\_454}

\begin{document}

\title{IDE-3D: Interactive Disentangled Editing for High-Resolution 3D-aware Portrait Synthesis}

\author{Jingxiang Sun}
\affiliation{%
  \institution{Tsinghua University}
  \country{China}
}

\author{Xuan Wang}
\affiliation{%
  \institution{Tencent AI Lab}
  \country{China}}

\author{Yichun Shi}
\affiliation{%
  \institution{ByteDance Inc.}
  \country{USA}
}

\author{Lizhen Wang}
\affiliation{%
 \institution{Tsinghua University}
 \country{China}}

\author{Jue Wang}
\affiliation{%
  \institution{Tencent AI Lab}
  \country{China}}

\author{Yebin Liu}
\affiliation{%
  \institution{Tsinghua University}
  \country{China}
}



\renewcommand{\shortauthors}{Sun et al.}

\begin{abstract}
Existing 3D-aware facial generation methods face a dilemma in quality versus editability: they either generate editable results in low resolution, or high quality ones with no editing flexibility. In this work, we propose a new approach that brings the best of both worlds together. Our system consists of three major components: (1) a 3D-semantics-aware generative model 
that produces view-consistent, disentangled face images and semantic masks; (2) a hybrid GAN inversion approach that initialize the latent codes from the semantic and texture encoder, and further optimized them for faithful reconstruction; and (3) a canonical editor that enables efficient manipulation of semantic masks in canonical view and producs high quality editing results. Our approach is competent for many applications, e.g. free-view face drawing, editing and style control. Both quantitative and qualitative results show that our method reaches the state-of-the-art in terms of photorealism, faithfulness and efficiency.
\end{abstract}



\begin{teaserfigure}
\centering
  \includegraphics[width=0.95\textwidth]{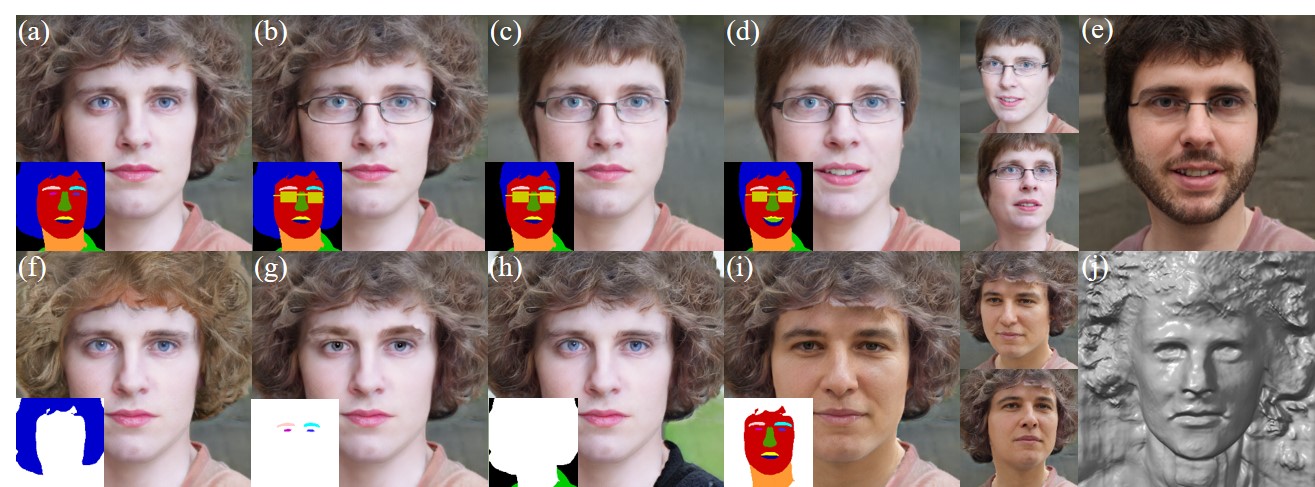}
  \caption{Our 3D portrait image generator allows users to perform interactive global and local editing on shape and texture in a view-consistent way. First row: starting from the source image (a), we gradually change the person's glasses (b), haircut (c), expression (d) and the global texture (e) with view consistency. Second row: Our method allows for disentangled region-level texture adjustment including hair (f), eyes \& eye brows (g), cloth \& background (h), the whole face region (i), etc.. Moreover, our method learns high-quality geometry from a collection of 2D images without multi-view supervision (j).}
  \Description{Enjoying the baseball game from the third-base
  seats. Ichiro Suzuki preparing to bat.}
  \label{fig:teaser}
\end{teaserfigure}

\maketitle

\section{Introduction}

Portrait synthesis finds broad applications in computer graphics including virtual reality (VR), augmented reality (AR), avatars-based telecommunication and so on. In this context, a generator is expected to synthesize photo-realistic facial images with control over multiple factors (e.g. haircut, wearings, poses, pupil color). Existing StyleGAN-based approaches~\cite{abdal2021styleflow, shen2020interfacegan, chen2016infogan, tewari2020stylerig} achieve the editing ability by either learning attribute-specific directions in the latent space~\cite{abdal2021styleflow, shen2020interfacegan, chen2016infogan, tewari2020stylerig}, or learning a more disentangled and controllable latent space through conditioning various priors~\cite{zhu2020sean, wang2018high, park2019semantic, chen2020sofgan, shi2021semanticstylegan, chen2020deepfacedrawing, chen2021deepfaceediting, li2020deepfacepencil, tewari2020stylerig, deng2020disentangled, wang2021cross}. Such methods are effective on synthesizing 2D images, but suffer from view-inconsistency if directly applied for editing different views of a 3D face.

Exploiting implicit neural representations to build 3D-aware GANs has recently gained considerable attention. Early NeRF-based generators~\cite{chan2021pi,schwarz2020graf} are proposed to generate view-consistent portraits relying on volumetric representation, which is memory-inefficient and can only synthesize images with limits resolution and fidelity. Several approaches are proposed to alleviate this problem by using the CNN-based up-sampler \cite{gu2021stylenerf, eg3d, xue2022giraffe, niemeyer2021giraffe,xu2021volumegan,orel2021stylesdf}. Recently, Sun et al.~\cite{sun2021fenerf} attempt to enable local face editing in the generated NeRF representations via GAN inversion, however the results exhibit inadequate photorealism. Furthermore, the optimization-based GAN inversion is very time-consuming, thus is not suitable for real-time interactive editing tasks.

In this work, we propose a high-resolution 3D-aware generative model that not only enables local control of the facial shape and texture, but also supports real-time, interactive editing. Our framework consists of a multi-head StyleGAN2 feature generator, a neural volume renderer and a 2D CNN-based up-sampler. To disentangle different facial attributes, shape and texture codes are injected separately into both the shallow and the deep layers of the StyleGAN-based feature generator. The output features are then used to construct the spatially aligned 3D volumes of shape (encoded in facial semantics) and texture in an efficient tri-plane representation. Given the generated 3D volumes, free-view portraits can be rendered via the volume rendering and a 2D CNN-based up-sampler with satisfactory view-consistency and photorealism. 

To enable various face editing applications, it is necessary to map the input image and semantic mask to the latent space and edit the encoded face via GAN inversion techniques. One possible solution is to use optimization-based GAN inversion as done in FENeRF~\cite{sun2021fenerf}, whereas two obvious drawbacks exist. On the one hand, it is difficult to obtain a proper initialization that can produce optimal latent codes. On the other hand, optimization is usually too time-consuming for interactive editing. We thus adopt a hybrid GAN inversion approach. Given the input facial image and corresponding semantic masks, we use the texture and semantic encoders to obtain corresponding latent codes, which are used as the initialization for the optimization-based pivotal tuning~\cite{roich2021pivotal} to obtain high-fidelity reconstruction. 


Given the optimized latent codes, editing can be performed by directly drawing on the inverted semantic masks. These modifications can then be mapped into the latent space via the semantic encoders. Unfortunately, the output latent code of the encoders cannot be used to faithfully reconstruct the input images and semantic masks. This is because it is too difficult for these encoders to learn the mapping from pose-entangled (semantic) image space to the pose-invariant latent space.
It is possible to use the proposed hybrid GAN inversion again to improve faithfulness, however it makes the editing operation time-consuming. To overcome this limitation, we present a canonical editor: the additional encoders that always take as input the images in canonical view and map them into the latent space. Note that the optimization-based pivotal tuning runs only once per input image to get the inverted latent codes. We then perform all the following operations through the proposed canonical editor. Benefiting from view normalization, the editor then enables real-time editing without sacrificing faithfulness.

In summary, we propose a locally disentangled, semantics-aware 3D face generator which supports interactive 3D face synthesis and local editing.
Our method supports various free-view portrait editing tasks with the state-of-the-art performance in photorealism and efficiency.
Our main technical contributions are as followings:
\begin{itemize}
    \item We present a high-resolution semantic-aware 3D generator which enables disentangled control over local shape and texture.
    
    \item We propose an hybrid GAN inversion method, which can faithfully invert the facial image and semantic mask into the latent spaces.
    
    \item We propose a canonical editor module that is capable of real-time editing on free-view portrait with high quality results. 
\end{itemize}

\section{Related Work}
\subsection{Generative 3D-aware Image Synthesis} Generative adversarial networks have recently achieved tremendous breakthroughs in terms of image quality and editability for 2D image synthesis. In recent years, neural scene representations have lifted image generation into 3D settings with explicit camera control. Early voxel-based methods \cite{gadelha20173d, henzler2019escaping, nguyen2019hologan, nguyen2020blockgan, zhu2018visual} fail to synthesize the complex scenes and photo-realistic details due to the limited grid resolutions. The Implicit Neural Representations (INR), especially the Neural Radiance Fields (NeRF), which have proven to generate high-fidelity results in novel view synthesis, are introduced to 3D-aware generative models \cite{chan2021pi, schwarz2020graf}. However, the costly sampling and MLP queries in neural volume rendering make those unsuitable for training high-resolution GANs. To this end, the recently developed 3D-aware GANs make use of a two-stage rendering process \cite{xue2022giraffe, niemeyer2021giraffe, gu2021stylenerf, orel2021stylesdf, eg3d, xu2021volumegan} or efficient sampling strategy \cite{zhou2021cips3d, deng2021gram} to tackle this problem. Meanwhile, aimed to reduce view-inconsistent artifacts brought by the 2D renderers, they adopt different strategies including NeRF path regularization~\cite{gu2021stylenerf}, dual discriminators~\cite{eg3d}, etc. Based on 3D-aware GANs, FENeRF~\cite{sun2021fenerf} has attempted to edit the local shape and texture in a facial volume via GAN inversion. Nonetheless, this method suffers from the low-resolution results and its optimization-based inversion cannot meet the demand for real-time and user-interactive applications.

\subsection{Conditional Portrait Editing} GANs are widely leveraged to learn a mapping from a reference in source domain to the target domain. Specifically for portrait synthesis, the reference are often referred to semantic masks~\cite{park2019semantic,zhu2020sean}, or hand-written sketches~\cite{chen2021deepfaceediting,chen2020deepfacedrawing}. In the context of semantic-guided facial image synthesis, SPADE~\cite{park2019semantic} leverages the semantic information to modulate the image decoder for better visual fidelity and texture-semantic alignment. Zhu et al.~\cite{zhu2020sean} further enable the region-wise control over the texture style relying on its region-adaptive normalization. Recently, SofGAN~\cite{chen2020sofgan} has been proposed to use semantic volumes for 3D editable image synthesis. Nevertheless, it requires multi-view images and the ground-truth 3D geometry to learn the semantic volumes. In addition, it lacks any mechanism for preserving the view-consistency in the synthesized textures.

\begin{figure*}[htbp]
  \centering
  \includegraphics[width=0.95\textwidth]{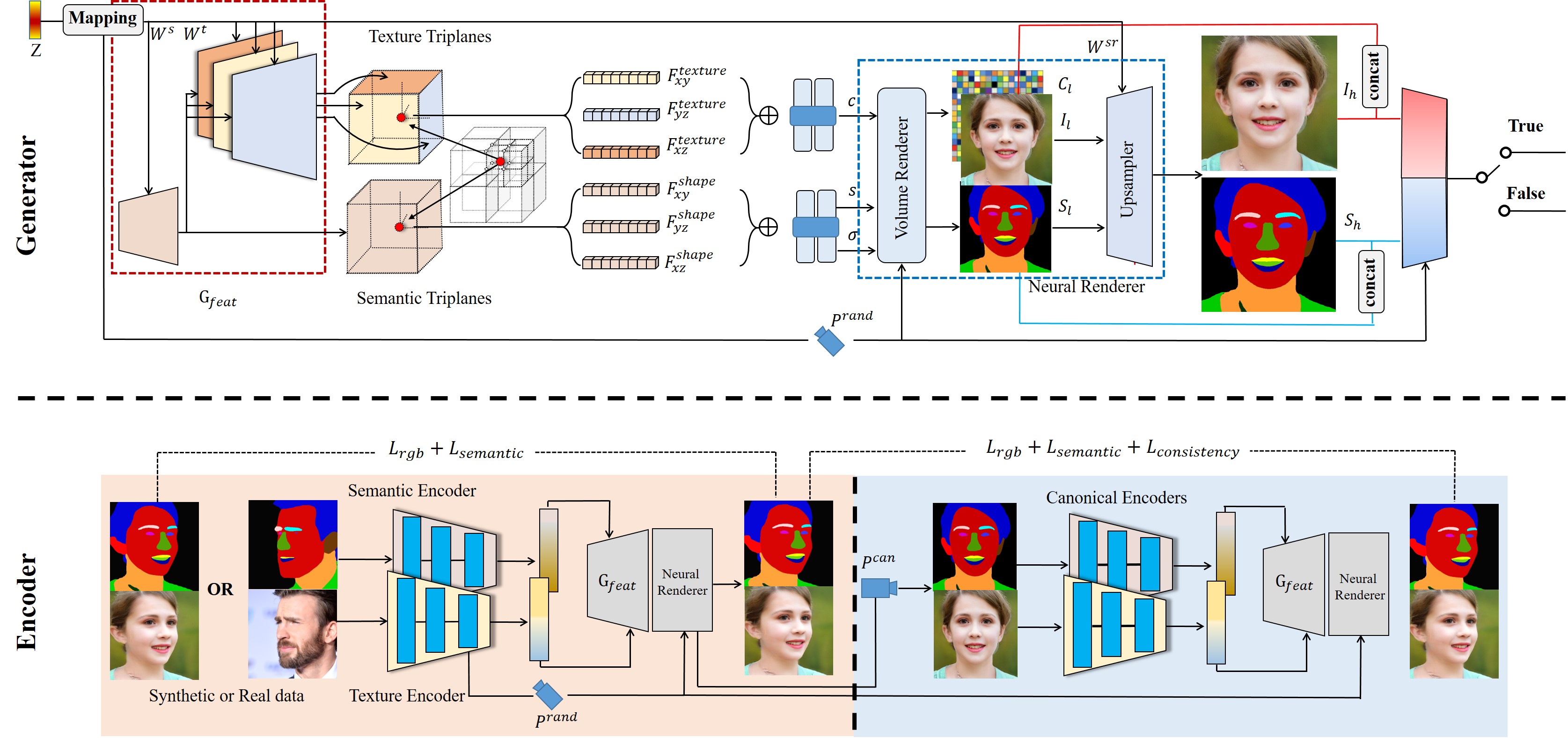}
  \caption{Pipeline of our 3D generator and encoders. The 3D generator (upper) consists of several parts. First, a StyleGAN feature generator $\mathrm{G}_{feat}$ constructs the spatially aligned 3D volumes of semantic and texture in an efficient tri-plane representation. To decouple different facial attributes, shape and texture codes are injected separately into both the shallow and the deep layers of $\mathrm{G}_{feat}$. Moreover, the deep layers are designed to three parallel branch corresponding to each feature plane to reduce the entanglement among them. Given the generated 3D volumes, RGB images and semantic masks can be rendered jointly via the volume rendering and a 2D CNN-based up-sampler. Encoders (lower) embeds the portrait images and corresponding semantic masks into the texture and semantic latent codes by two independent proposed encoders. With a predicted camera pose, Then the fixed generator reconstructs the portrait under the predicted camera pose. In order to eliminate pose effect, we jointly train a canonical editor which takes as input the portrait images and semantic masks under the canonical view, with the consistency enforcement.}
  \Description{}
  \label{pipeline}
\end{figure*}

\subsection{GAN Inversion}

GAN inversion techniques play an important role for editable portrait synthesis. Existing methods can be roughly categorized into three classes: optimization-based, learning-based and hybrid GAN inversion approaches. Optimization methods~\cite{abdal2019image2stylegan,abdal2020image2stylegan++} directly optimize latent codes by minimizing the difference between the input image and the reconstructed ones, which are usually harmed by the improper initialization and low efficiency. By contrast, learning-based methods exploit an encoder network to directly map the input images into the latent space. Richardson et al.~\cite{richardson2021encoding} first present to train an encoder to map the images into the $W^{+}$ space. Alaluf et al.~\cite{alaluf2021restyle} further propose iterative forward refinement to predict latent codes with better image preservation. Most recently, Wang et al.~\cite{wang2021HFGI} propose the distortion consultation branch to recover high-frequency image-specific details. In the case of hybrid GAN inversion, it leverages the encoder to predict a latent code which is used as the initialization in the optimization process~\cite{zhu2020indomain}. In addition to optimizing the latent codes, Pivotal Tuning Inversion~\cite{roich2021pivotal} also fine-tunes the generator parameters for recovering image details that cannot be encoded in the latent space.


\section{Method}

In this section, we provide a more detailed description of the presented approach, IDE-3D. The organization of this section is as follows: We introduce the 3D-semantic-aware generator in Sec.~\ref{sec:3.1}, including the network architecture and the training losses. In Sec.~\ref{sec:3.2}, we first describe the encoders employed in then hybrid GAN inversion. Then we present the design of the canonical editor, as well as explain the training loss and strategy. 

\subsection{Photorealistic 3D-Semantic-Aware GAN}
\label{sec:3.1}

In order to synthesize the high-definition portrait image with flexible editability, as shown in~\ref{pipeline}, our generator consists of the Multi-head StyleGAN-based feature generator which generates the semantic-disentangled 3D representations and the neural renderer that is composed of the volume renderer and a 2D CNN-based up-sampler.


\noindent \textbf{Multi-head StyleGAN generator.} 
We construct the generator with a StyleGAN-based backbone and dual Multi-Layer Perceptron (MLP) decoders.
Given a randomly sampled latent code $z\in \mathbb{R}^{512}$ and camera pose $p\in \mathbb{R}^{25}$, we first map $z$ to the $\mathcal{W}_+$ space via the network $\mathrm{M}$. Different from EG3D~\citet{eg3d}, we inject the latent codes $w_s$ and $w_t$ into the shallow (first eight layers) and the deep (last ten layers) layers of the StyleGAN generator $\mathrm{g}$ respectively to obtain the disentangled semantic $f^{s}_{triplane}$ and texture $f^{t}_{triplane}$ features in tri-plane formulation.


In particular, we introduce a multi-head architecture. The deep layers of StyleGAN-based backbone are split into three branches which are connected to the same shallow layers. In other words, the three axis-aligned orthogonal texture feature planes $f^{t}_{triplane}$ are generated from different branches individually. We empirically find that the benefit of the multi-head architecture is two-head. On one head, it facilitates training convergence. On the other hand, it helps to avoid the "hollow face illusions", especially for the illusion of the eyes that follow the viewing camera.

\noindent \textbf{Dual MLP decoders.} 
Our Dual encoder $\mathrm{\Phi}$ has two independent modules $\mathrm{\Phi}_{s}, \mathrm{\Phi}_{t}$, for facial semantic and texture, both of which are formulated as a single hidden layer MLP of 64 units.
For each point $x$ in the 3D space, we query its semantic and texture feature vectors $f^{s}_{x}, f^{t}_{x}$ by projecting it to axis-aligned planes of $f^{s}_{triplane}$ and $f^{t}_{triplane}$. The $\mathrm{\Phi}_{s}$ decodes the semantic features $f^{s}_{x}$ to the volume density $\mathbf{\sigma}$ and semantic labels $\mathbf{s}$. And the multi-channel colour feature $c$ is decoded  by $\mathrm{\Phi}_{t}$ from $f^{t}_{x}$. The total function of our semantic-aware decoder $\mathrm{\Phi}$ can be written as below:

\begin{equation}
    \begin{split}
     \mathrm{\Phi}: 
     (x, f^{s}_{x}, f^{t}_{x}) \mapsto (\sigma, c, s).
    \end{split}
\end{equation}

\noindent \textbf{Neural renderer.} We adopt a neural volume renderer and a 2D CNN-based up-sampler to render the portrait image from the decoded 3D volume. Given the all the sampled volume densities, semantic labels and color features of all the pixel rays $r$, we exploit the volume renderer to obtain a colour feature map $C_l$ and a semantic mask $S_l$ in low resolution. In specific, the first three channels of $C_l$ store the $(r,g,b)$ values. It contains a low-resolution portrait in the color feature map.



In order to bring the image rendering into high resolution and photorealism, a 2D CNN-based up-sampler is introduced. The rendered intermediate color feature $C_l$ and semantic map $S_l$ are up-sampled into the final high-resolution RGB image $I_h$ and semantic map $S_h$ via a two StyleGAN2 synthesis blocks which are modulated by the corresponding latent codes and predict the residuals of the LR inputs.

\noindent \textbf{Dual discrimination and training objective.} Aimed to model the joint distribution of aligned real RGB images and semantic masks $p(I, S)$, we present the dual discriminator $\mathrm{D}_{dual}$ which takes as input both RGB images and semantic masks. As mentioned before, the first three channels of feature map $C_l$ exactly form a low-resolution RGB image.  The dual discriminator thus takes as input the concatenation of resized LR rgb/semantics image and the synthesised HR rgb/semantic image $(I_h, I_l', S_h, S_l')$.

$\mathrm{D}_{dual}$ is designed with has two independent convolution blocks as well as minibatch standard deviation layers for RGB images $(I_h, I_l')$ and semantic m$(S_h, S_l')$, followed by a concatenation before the final output. The dual-branch discriminator has two advantages: 1) Decoupling RGB image and semantic mask reduces the unnatural contours of synthesized images effected by semantic mask; 2) It supports for regularizing the gradient penalty of image and semantic branches separately. We obtain the discrimination loss as follows: 
\begin{equation}
    \begin{split}
    \mathcal{L}_{\mathrm{D}_{dual}, \mathrm{G}} = 
    &\mathbb{E}_{z \sim \mathcal{Z}, p \sim \mathcal{P}}[f(\mathrm{D}_{dual}(\mathrm{G}(z, p)))] + \\
    &\mathbb{E}_{I \sim \mathcal{P_{I}, S \sim \mathcal{P_{S}}}}[f(-\mathrm{D}_{dual}(I, I_r, S_r, S) + \\
    &\lambda_{image}\|\mathbf{\bigtriangledown}\mathrm{D}_{dual}(I)\|^2) + \lambda_{seg}\|\mathbf{\bigtriangledown}\mathrm{D}_{dual}(S)\|^2)]
    \end{split}
    \label{eq:generator loss}
\end{equation}

\noindent \textbf{Density regularization} As pointed in EG3D~\cite{eg3d}, we sometimes observe the "seam" artifacts along the edge of the faces in both synthesized image and the extracted mesh. A possible reason could be the imbalanced camera pose distribution of datasets. In particular, the distribution of head pose angles, in FFHQ dataset, has relatively small variance, thus it is challenging to learn a good shape of faces rendered from side views. In IDE-3D, we introduce a density regularization loss which regularize the density field such that nearby 3D points have similar density values, inspired by \cite{eg3d, oechsle2021unisurf, chen2021snarf}. The density regularization loss is:

\begin{equation}
    \begin{split}
    \mathcal{L}_{density} = 
    &\mathbf{\sum}_{x_{s} \in \mathcal{S}}\|d(x_{s}) - d(x_{s}+\epsilon)\|_2
    \end{split}
    \label{eq:generator loss}
\end{equation}

\noindent Here, $\mathcal{S}$ denotes the set of fine points, $d(x)$ is the predicted density of point $x$. The total learning objective is:

\begin{equation}
    \begin{split}
    \mathcal{L}_{total} = 
    &\mathcal{L}_{\mathrm{D}_{dual}, \mathrm{G}} + \lambda_{density} \mathcal{L}_{density}
    \end{split}
    \label{eq:generator loss}
\end{equation}

\subsection{Hybrid GAN Inversion and Canonical Editor}
\label{sec:3.2}


Given a pre-trained generator, the GAN inversion approach is necessary for interactive editing. In this context, photorealism, faithfulness and efficiency are the most important factors.

\noindent \textbf{Encoder training and multi-view augmentation.}
A hybrid GAN inversion method, which can be regarded as a combination of the learning-based and optimization-based approaches, is introduced. As illustrated in Fig.~\ref{pipeline}, we adopt two encoders that map the input image and semantic map to the latent spaces $\mathcal{W}^{s}_+$ and $\mathcal{W}^{t}_+$. In our case, the camera pose is also needed to be extracted from the pose-entangled 2D images. We achieve it by appending two dimensions in the output of the texture branch for predicting yaw and pitch. We adopt two types data during training. One is the real data from specific dataset. The other is synthetic which can be rendered by the pre-trained generator. Furthermore, we present a multi-view augmentation strategy. In particular, for a synthetic portrait image, we additionally render it from another $k$ random views and train them together with two-fold consistency constrains. On the one hand, as they are synthesized from the same latent code, their inverted codes should be the same. On the other hand we also add a high-level feature consistency loss on the reconstructed images. We observe that the multi-view augmentation benefits a better content-pose disentanglement and improve the stability of learning-based inversion across different views. During real portrait editing, we gain the inverted latent code as an initialization for pivotal tuning \cite{roich2021pivotal} to faithfully reconstruct the inputs.

\noindent \textbf{Canonical editor.} 
In conjunction with the optimization-based tuning, the presented hybrid GAN inversion can reconstruct the inputs faithfully. Unfortunately, the time-consuming tuning process is impractical for the real-time tasks. The possible reason that the encoders cannot yield the faithful latent codes is learning the mapping from pose-entangled images to the pose-invariant latent space could be too difficult. To tackle this problem, we propose a specific encoder, the canonical editor $E_{can}$ which only map canonical-viewed images and semantic masks into latent spaces as shown in bottom-right of Fig.~\ref{pipeline}. $E_{can}$ takes as input canonical-view synthetic face images and semantic masks rendered from a pretrained generator in training. During real portrait editing, we gain the reconstructed latent code by exploit the aforementioned hybrid inversion only once, and all the following editing operations are performed thorough the $E_{can}$. Experiments demonstrate that the canonical editor produces much more faithful results compared with the other encoders.


\begin{figure}[t]
  \centering
  \includegraphics[width=\linewidth]{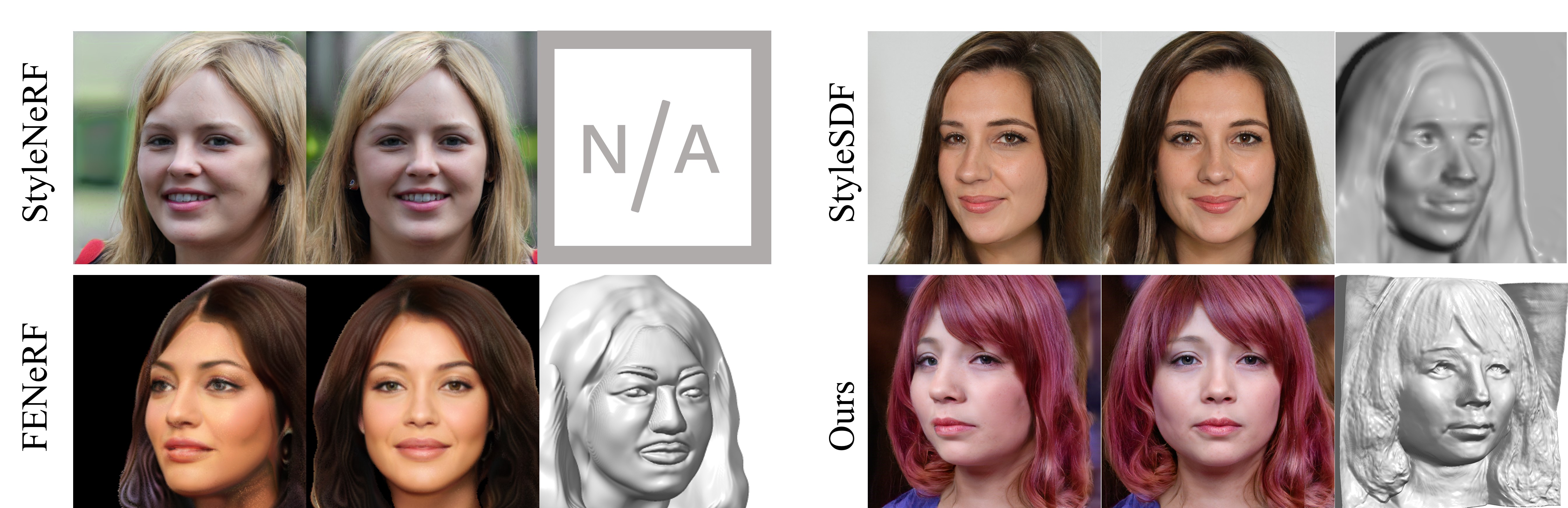}
  \caption{Qualitative comparison among FENeRF, StyleNeRF, StyleSDF and ours. Models are trained with FFHQ at 512$^{2}$, except FENeRF with FFHQ at 128$^{2}$. Note that we don't extract shape of StyleNeRF as there is no released scripts in their released codes.}
  \Description{}
  \label{comparison}
\end{figure}

\section{Applications and Experiments}
In this section we conduct a series of qualitative and quantitative experiments to demonstrate the superiority of our framework from three dimensions: photorealism, faithfulness and editability \cite{xia2021gan}. Sec.~\ref{sec:4.1} illustrates that our framework gets comparable performance on photorealism as EG3D which is the current SOTA; Sec.~\ref{sec:4.2} proves the effectiveness of our proposed strategies on the faithfulness of inversion; Sec.~\ref{sec:4.3} demonstrates various applications by exploring the editability of global and local styles.

\noindent \textbf{Datasets.}  We train and evaluate our models on two popular datasets for face synthesis and editing: CelebAHQ-Mask\cite{CelebAMask-HQ} and FFHQ\cite{karras2019style}. For both datasets, we predict approximate camera extrinsics for each image with a off-the-shelf head pose estimator\cite{deng2019accurate}, and all images share a set of constant camera intrinsics. For semantic masks, we adopt the provided ones of CelebAHQ-Mask and label the semantic classes for FFHQ with a pretrained face parser following SofGAN. For data augmentation, we augment both datasets by horizontal flip as well as rebalance which is aimed to increase the sampling probability of images photoed at steep angles.

\begin{table}[t]
\centering
\begin{tabular}{l|lll|lll} 
\hline
                   & \multicolumn{3}{c}{FFHQ}                                                                                   \vline  & \multicolumn{3}{c}{CelebAHQ-Mask}                                                                             \\
                   & \multicolumn{1}{c}{FID$\downarrow$} & \multicolumn{1}{c}{KID$\downarrow$} & \multicolumn{1}{c}{ID$\uparrow$}
                   \vline & \multicolumn{1}{c}{FID$\downarrow$} & \multicolumn{1}{c}{KID$\downarrow$} & \multicolumn{1}{c}{ID$\uparrow$}  \\ 
\hline
FENeRF~$128^2$      & 29.0                                & 3.728                                & 0.61                                 & 13.2                                & 1.6                                 & 0.68                                  \\
StyleNeRF~$512^{2}$ & 7.8                                 & 0.22                                 &  0.62                                & 16.6                                   & 1.129                                    & 0.59                                 \\
StyleSDF~$512^{2}$  & 13.1                                & 0.269                                &  0.742                                & 7.28                                    &  0.280                                   &  0.65                                 \\
EG3D*~$512^{2}$      & 4.7                                 & 0.132                                    &  0.77                                & --                                    & --                                    & --                            \\

Ours~$512^{2}$     & 4.6                                 &       0.130                               &      0.76                            & 4.9                                     & 0.135                                    & 0.75                                  \\
\hline
\end{tabular}
  \caption{Quantitative comparison with FENeRF, StyleNeRF, StyleSDF and EG3D on FFHQ and CelebAHQ-Mask dataset. * stands for quoting from the paper.}
  \label{tab:comparison}
\end{table}

\subsection{Comparisons}
\label{sec:4.1}
\noindent \textbf{Baselines.} We select different baselines for various tasks. For image synthesis and geometry quality, we compare our method with the state-of-the-art high-resolution 3D generative models, StyleNeRF, StyleSDF and EG3D, as well as FENeRF which share some similarity with us on semantic awareness and editing ability on local shape. For the task of semantic-guided synthesis and editing, we compare our method with FENeRF and 2D GANs, including SPADE, SEAN and SofGAN. 

\begin{figure}[b]
  \centering
  \includegraphics[width=\linewidth]{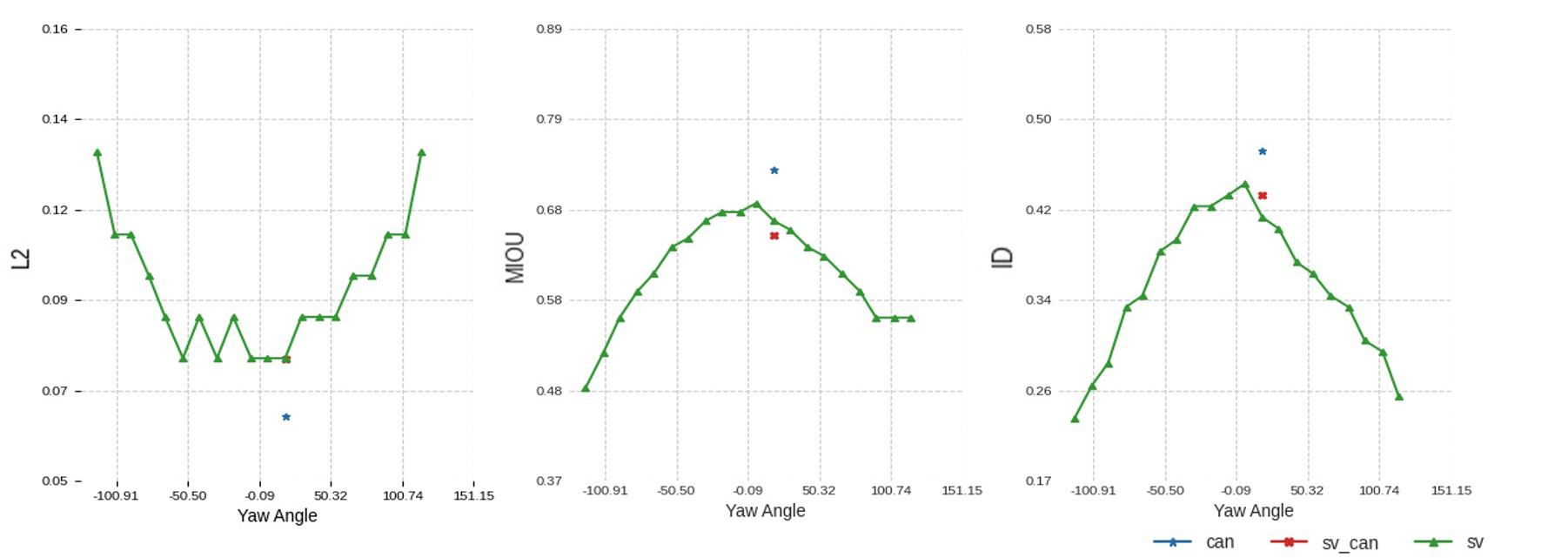}
  \caption{The inversion ability of T/S encoder degenerates along with the increasing yaw angles. Input with the fixed canonical view (green "X" point) helps to stabilize the performance. Our proposed canonical editor (blue "star" point) shows the superior inversion capacity, indicating the effectiveness of learning style mapping from the canonical view.}
  \Description{}
  \label{fig:chart}
\end{figure}

\noindent \textbf{Qualitative comparisons.} Fig.~\ref{comparison} presents a qualitative comparison against baselines. FENeRF synthesizes view-consistent images, while the rendering resolution and image quality are limited. StyleNeRF and StyleSDF show impressive image synthesis quality and StyleSDF also learns a good quality shape with the implicit SDF representation and geometry constraint during training, but it fails to capture shape details. our method achieves not only higher photo-realism in image synthesis but also higher-fidelity geometry.

\noindent \textbf{Quantitative evaluations.} We quantitatively evaluate the photo-realism and the diversity of image synthesis using he Frechet Inception Distance (FID) \cite{heusel2017gans} and Kernel Inception Distance (KID) \cite{binkowski2018demystifying}. Following \cite{eg3d}, we evaluate view consistency assessed by multi-view facial identity consistency (ID). This metric calculates the mean Arcface [9] cosine similarity score between pairs of the same synthetic face rendered from two random camera poses. Table.~\ref{tab:comparison} reports the quantitative metrics comparing with baselines on the FFHQ and CelebAHQ-Mask datasets. Our method shows the state-of-the-art image synthesis quality and view consistency across all datasets. 

\subsection{Ablation Study}
\label{sec:4.2}

\begin{figure}[t]
  \centering
  \includegraphics[width=\linewidth]{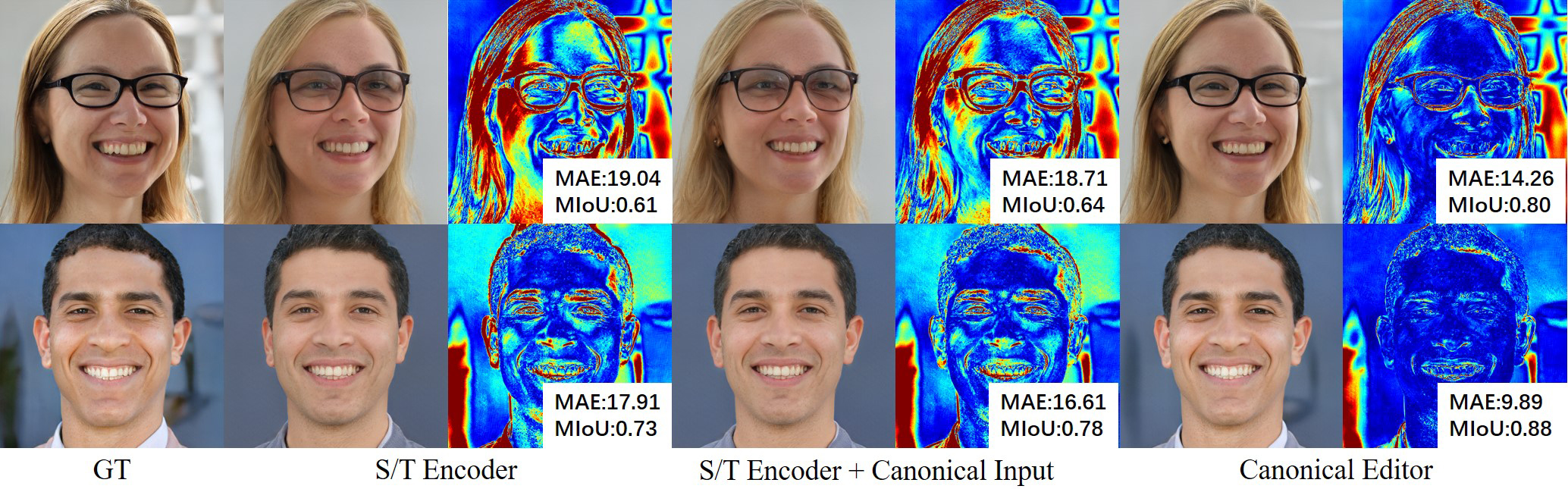}
  \caption{Effectiveness of our canonical style encoder. We compare the inversion results of the standard encoder, the standard encoder with canonical input and our canonical encoder. Both the qualitative results and the quantitative errors prove the superiority of our canonical style encoder.}
  \Description{}
  \label{fig:ablation_can}
\end{figure}

\noindent \textbf{Canonical editor.} Recall that we propose two encoders, a general encoder $\mathrm{E}$ and a canonical editor $\mathrm{E}_{can}$. In order to investigate the effectiveness of $\mathrm{E}_{can}$, we conduct experiments to explore: 1) how the input view angles influence on the inversion quality of $E$; 2) Is it easier to learn the style mapping under the canonical view than random views? First we sample 4000 camera poses using a stratified
sampling approach ranging from t -63$^{\circ}$ to +63$^{\circ}$. Then we sample 4000 latent codes and render each in a sampled view for $E$ input $I$, the front view for $\mathrm{E}_{can}$ input $I_{can}$, and a random view for test $I_{test}$. Given the predicted latent code from $\mathrm{E}$ and $\mathrm{E}_{can}$, we utilize them to render the test view and compute the similarity with $I_{test}$, respectively. Furthermore, in order to eliminate the data effect, we also process the canonical view image $I_{can}$ into $\mathrm{E}$. Fig~\ref{fig:chart} plots the inversion performance of the three settings above assessed by 3 metrics against the yaw angles of our sampled views. We can see that the inversion faithfulness of $\mathrm{E}$ degenerates with the head pose getting far from the canonical view, which indicates the steep view angles truly affect the style mapping ability. Moreover, while the performance of $\mathrm{E}_{can}$ and $\mathrm{E}$ with the canonical input both remain stable along with the changing poses, $\mathrm{E}_{can}$ illustrates the superior performance on all metrics, indicating it is easier to learn a good style mapping in the canonical view. Fig~\ref{fig:ablation_can} shows two examples of our experiment results. 

\begin{figure}[t]
  \centering
  \includegraphics[width=\linewidth]{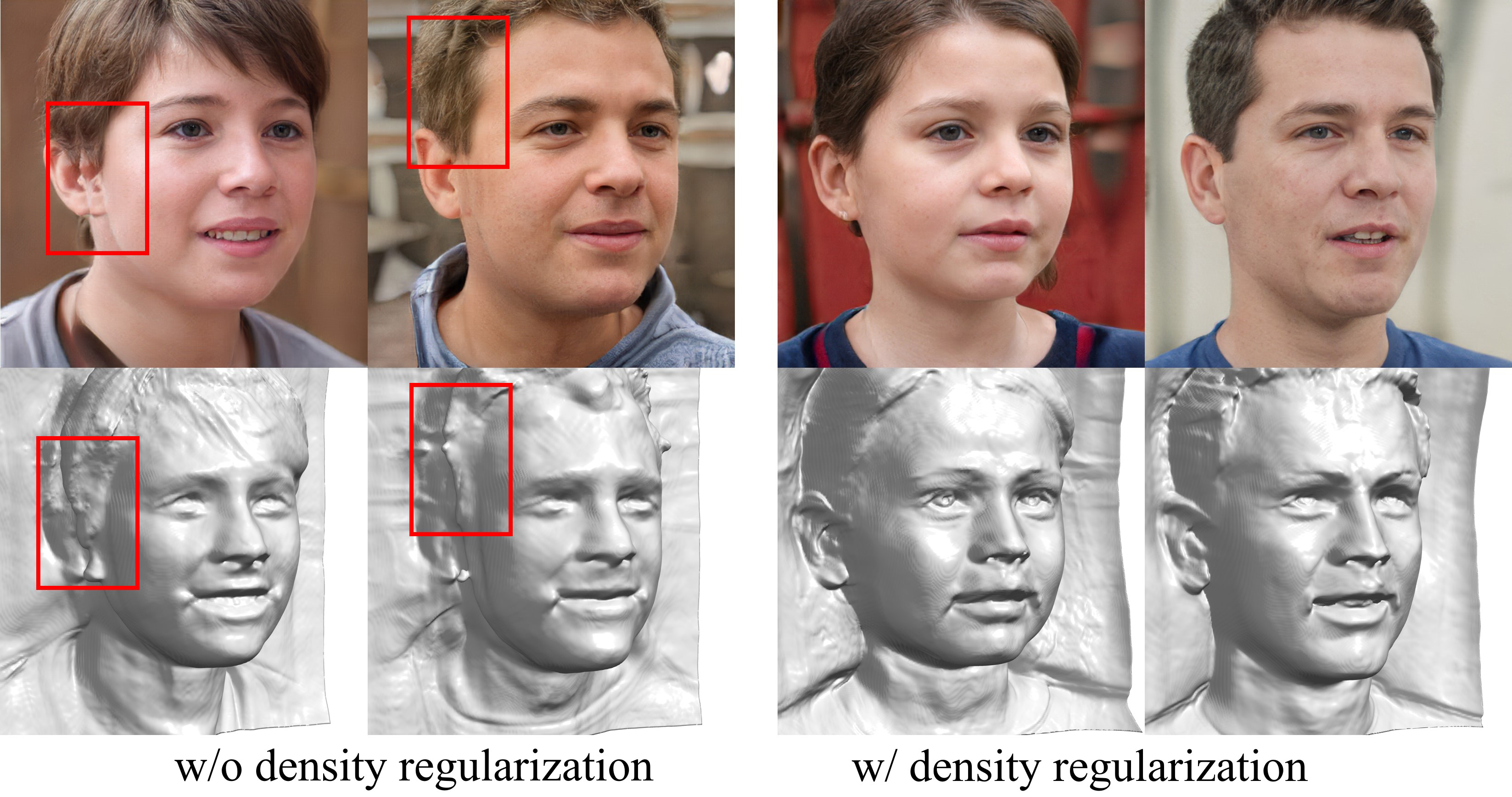}
  \caption{Effect of density regularization. We synthesize images and geometry for seeds 0-3 with truncation $\phi=0.5$, using two models w/ and w/o density regularization.}
  \Description{}
  \label{fig:density}
\end{figure}

\noindent \textbf{Density regularization.} In this work we propose the density regularization to enforce the nearby points to the similar volume density, which effectively remove the "seam" shape artifacts of side faces. Fig.~\ref{fig:density} demonstrates the comparison between our model trained w/ and w/o density regularization. It is obviously that the renderings of the model w/o density regularization show "seam" artifacts in both images and the extracted meshes, with varying degrees. On the contrary, density regularization wipe off such artifacts without any degeneration of shape quality. 

\begin{figure}[b]
  \centering
  \includegraphics[width=0.85\linewidth]{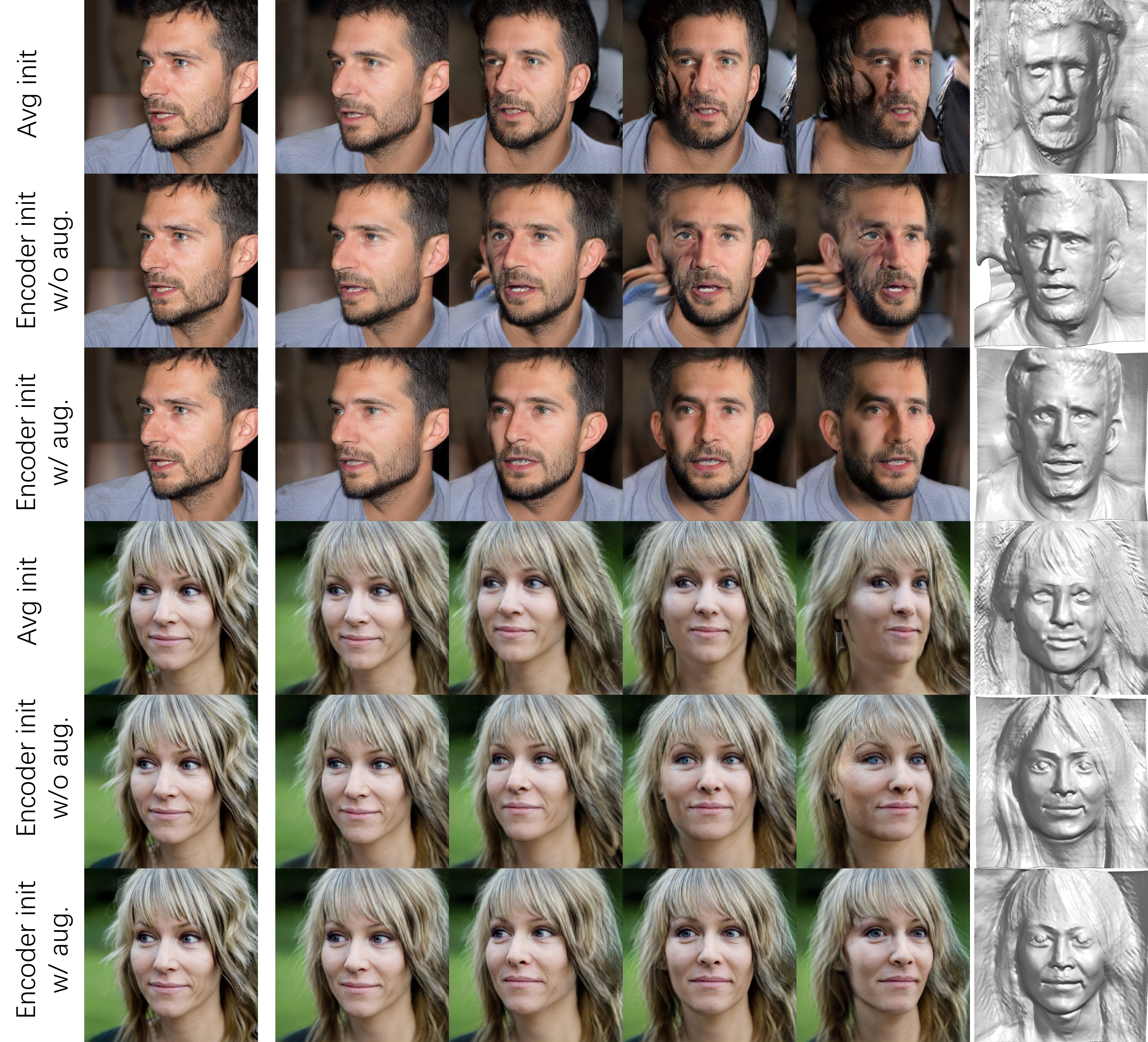}
  \caption{Different initialization strategies for hybrid GAN inversion. Multi-view augmentation helps to alleviate inconsistent artifacts within self occluded regions and reconstruct more faithful shapes. }
  \Description{}
  \label{fig:multiview}
\end{figure}

\begin{figure*}[t]
  \centering
  \includegraphics[width=0.95\textwidth]{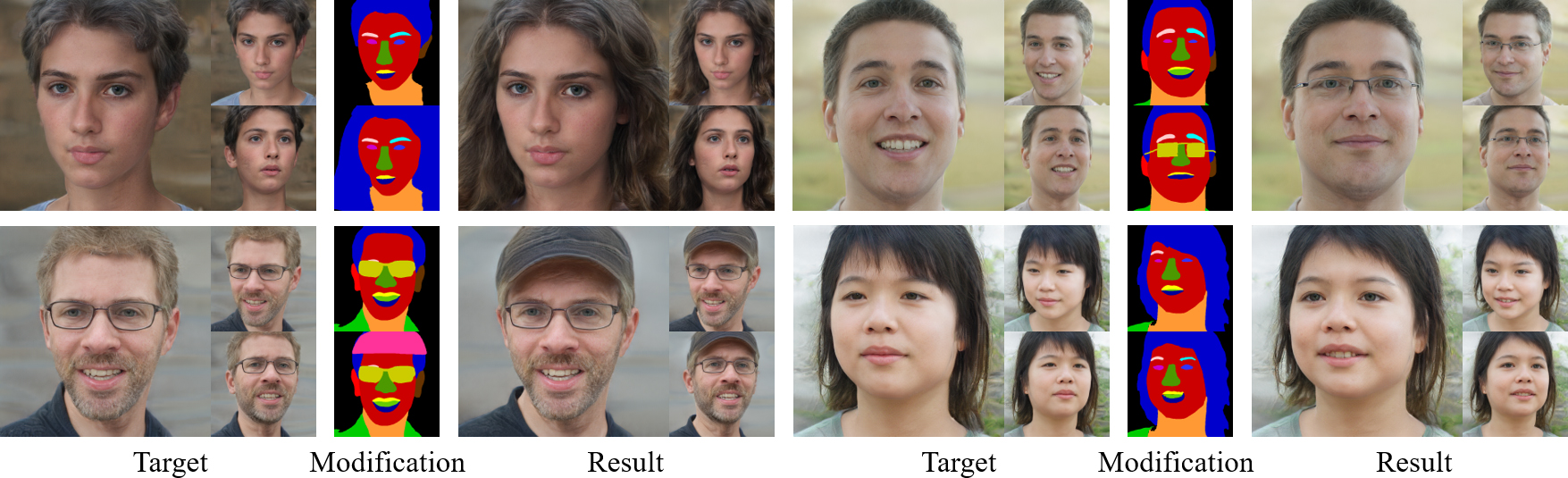}
  \caption{Interactive 3D face drawing. Our method supports interactively synthesize view-consistent photo-realistic portrait images by drawing a semantic mask. The figure shows some facial attribute editing examples, e.g. hairstyle, glass (the first raw), hat, eyes and expression (the second raw). We can observe that our method preserves the unedited local regions well when manipulating the shape and expressions.}
  \Description{}
  \label{fig:interactive}
\end{figure*}

\noindent \textbf{Hybrid GAN inversion.} We perform the experiments to illustrate the effectiveness of the proposed hybrid GAN inversion. Fig.~\ref{fig:multiview} demonstrates the inversion performance under three kinds of initialization approaches: the averaged random latent codes (Avg init), reconstructed latent code by the encoder with (without) multi-view augmentation (Encoder init w/ or w/o aug.). Note the first initialization manner is exactly the optimization-based inversion method. Recall the multi-view augmentation. During training encoders, we introduce multi-view augmentation, i.e. training with multi-view rendered synthetic data with consistency enforcement to encourage the coder to decouple style encoding from pose-entangled RGB (semantic) images.
We can see that the encoder without multi-view augmentation provides a reasonable shape and texture prior thus obtains more consistent faces under novel views than average initialization. However, it still suffers from inconsistent artifacts in novel views with self occlusion (the 5th column in both cases). On the contrary, the encoder with multi-view augmentation significantly alleviate the inconsistent artifacts and also learns a more faithful geometry,  indicating the augmentation helps to learns a more view-consistent style mapping.

\subsection{Applications}
\label{sec:4.3}
\noindent \textbf{Interactive neural 3D face drawing and editing.} IDE-3D supports interactive 3D face drawing and editing. Given a semantic mask taken from real portrait images or randomly sampled from the shape latent space, our method can generate a 3D face with the same layout as the input semantic mask. As demonstrated in Fig.~\ref{fig:interactive}, we interactively edit the semantic mask to obtain the user-desired shapes in a view consistent manner. 


\begin{figure}
  \centering
  \includegraphics[width=\linewidth]{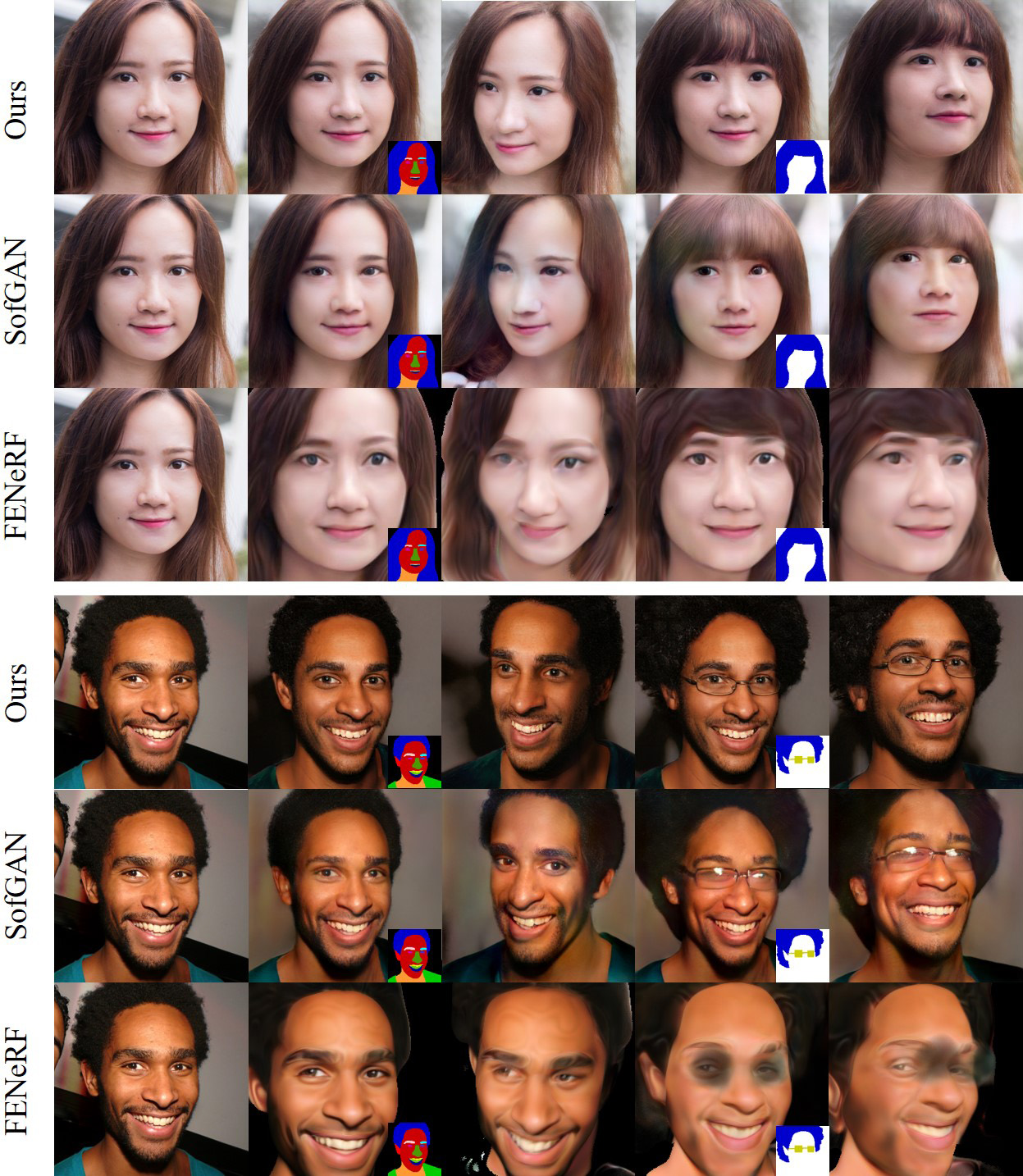}
  \caption{Real Portrait local shape editing. IDE-3D supports high-quality interactive real face editing using the semantic mask in a view consistent manner, which can not be guaranteed in SofGAN and FENeRF.}
  \Description{}
  \label{fig:real_edit}
\end{figure}

\noindent \textbf{Region-level texture editing.} Our semantic-aware synthesis framework enables region-level texture style controlling. Benefiting from the semantic segmentation branch of neural decoder, we can predict the semantic labels of each 3D point. By selecting user-specific target regions( e.g. hair, eyes, cloth etc), we can construct a 3D style mask which only select 3D points belonging to these semantic classes according to their semantic labels. Then given a desired texture style, we utilize it to generate new texture triplane and replace the color of selected 3D points by querying from this triplane. Fig.~\ref{fig:teaser} demonstrates the region-level texture style adjustment in hair, lips, cloth and background. Our approach is capable of maintaining the texture style unchanged except the target regions. Please refer to our supplementary video for more visualization examples. 
\begin{figure}
 \centering
 \includegraphics[width=0.95\linewidth]{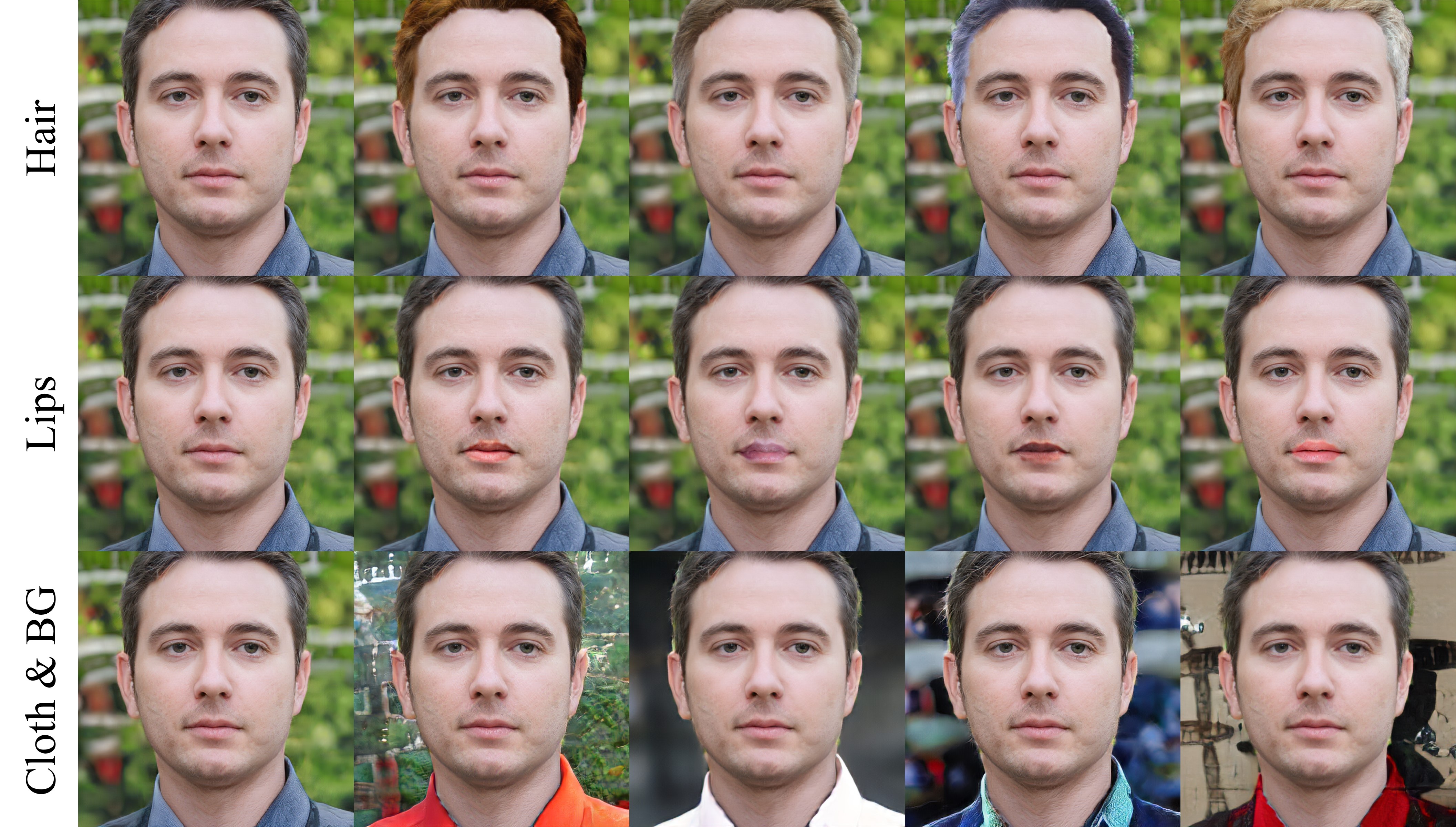}
 \caption{Region-level texture style adjustment. Our approach enables view-consistent region-level texture style adjustment on 19 classes: skin, eye glasses, eye brows, hair, nose, lips, cloth, wearings, etc.}
 \Description{}
 \label{texture editing}
\end{figure}

\noindent \textbf{Real Portrait local editing.}
Our IDE-3D enables high-fidelity 3D inversion and interactive editing of real portrait images. We take a comparison with SofGAN and FENeRF, two representative works in 2D and 3D semantic guided editing, as illustrated in Fig.~\ref{fig:real_edit}. First, We take three real portrait images and project them into the latent space of each synthesis framework. For a fair comparison, we adopt the unified optimization-based inversion method in StyleGAN2 \cite{karras2020analyzing} for all three frameworks. Turing to editing, we maintain their original editing methods for the best performance. Given an edited semantic mask under the input view, we combine it with the optimized latent code into a pretrained SofGAN model to obtain the edited portrait image. In order to obtain the free-viewed results, we need to project the semantic mask into its semantic field. Unfortunately, we fail to get a reasonable result. Therefore, we provide SofGAN with the free-viewed semantic masks rendered by our method instead. For FENeRF, we directly process the target semantic mask into it for the semantic inversion. Fig.~\ref{fig:real_edit} demonstrates that our method outperforms SofGAN and FENeRF in editing fidelity, image quality and view-consistency. SofGAN and our IDE-3D both obtain high-fidelity inversion result in the input view. However, SofGAN changes the identity obviously when turning to other views. FENeRF achieves high view consistency due its 3D nature, however, struggles to reconstruct high-fidelity texture detail. FENeRF also fails to model glasses due to its limited diversity of its GAN latent space. 

\section{Conclusion}
In this paper, we propose the IDE-3D consisting of a 3D-semantic-aware generator, a hybrid GAN inversion approach and the canonical editor. As demonstrated in massive experiments, our method supports many the flexible and interactive face editing tasks in real time without sacrificing faithfulness and photorealism.

\noindent \textbf{Limitations and future works.} As the free-view portrait is encoded in a 3D volume, using single image to reconstruct the 3D facial volume is an ill-posed problem. As a result, it could produce the in-plausible facial geometry in some cases, which is a known limitation  to our method. To this end, we plan to study the multi-view/frame GAN inversion approaches in future works.
\bibliographystyle{ACM-Reference-Format}
\bibliography{sample-base}

\appendix

\section{Overview}
In the supplement, we first discuss the implementation details of our framework (Sec.~\ref{sec:1}) including the network architecture and hyperparameters. We also provide the experiment details (Sec.~\ref{sec:2}) such as datasets and baselines. Finally, we provide additional visual results (Sec.~\ref{sec:3}). 
\section{Implementation details}
\label{sec:1}
We implement our framework on the top of the official PyTorch implementation of StyleGAN2, with the updated version: \href{https://github.com/NVlabs/stylegan3}{StyleGAN3}. We adopt the efficient synthesis structure and training designs of StyleGAN which incorporates equalized learning rate, label conditioned mapping network and discriminator, blur initialization tricks and GAN loss with R1 regularization. 

\noindent \textbf{Multi-head StyleGAN-based feature generator.} In the main paper we introduce a multi-head StyleGAN-based feature generator to disentangle feature planes in the texture tri-planes. In specific, the multi-head architecture is designed to the three parallel branches with a shared input feature map from the last layer of the semantic branches with dimension 64 $\times$ 64 $\times$ 64. Each texture branch upsamples the shared feature map into 256 $\times$ 256 $\times$ 32. 

\noindent \textbf{Semantic-aware neural renderer.} Our decoder consists of two independent multi-layer perceptrons (MLPs), each with a single hidden layer of 64 units and softplus activation functions. We let one decoder to predict both semantic labels $s \in \mathbb{R}^{19}$ and density $d \in \mathbb{R}^{1}$ as we observe that the 3D points belonging to the same semantic class are more likely to share the similar geometry. The other decoder is to predict a feature vector $c \in \mathbb{R}^{32}$. Following EG3D\cite{eg3d}, we treat the first three channels of $c$ as low-resolution RGB image $I_{l}$. 

\begin{figure}[b]
  \centering
  \includegraphics[width=\linewidth]{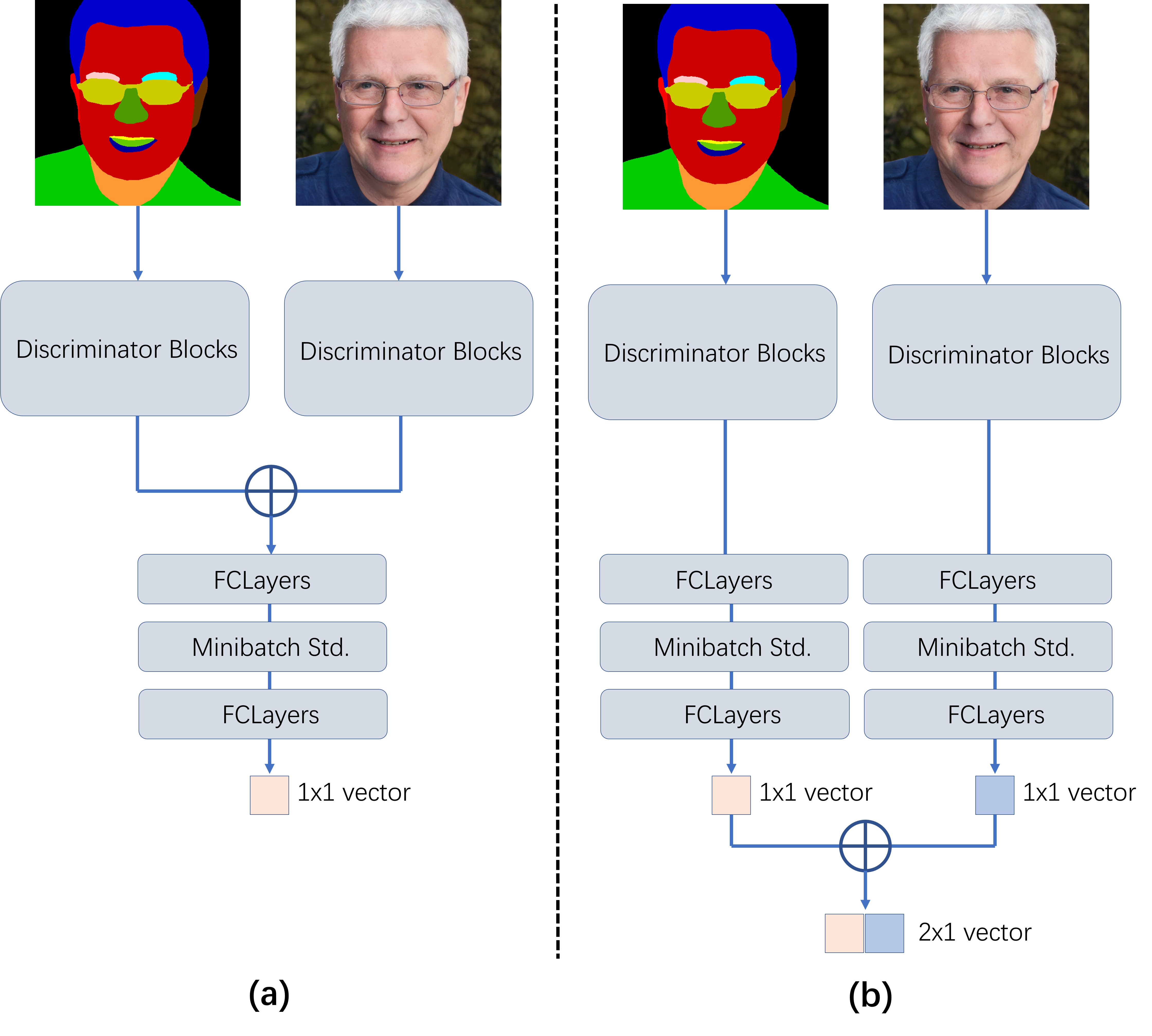}
  \caption{}
  \Description{}
  \label{fig:dual}
\end{figure}

\noindent \textbf{Dual-branch discriminator.} 
We introduce the dual branch discriminator to model the joint distribution of RGB images and semantic masks. Besides the naive concatenation mentioned in the main paper, we tried two dual branch approaches as illustrated in Fig.~\ref{fig:dual}. The first method (Fig.~\ref{fig:dual} a ) first processes semantic masks and RGB images into two independent discriminator blocks and concatenate them together to calculate the standard deviation. The output of the first method is 1 $\times$ 1 vector. Instead of concatenation, the second method shown in Fig.~\ref{fig:dual} b directly outputs the two vector independently for the semantic masks and RGB images. Though the second method doesn't explicitly enforce the semantic alignment, we experimentally found that it leads to a better image quality without lost semantic-texture consistency. 

\noindent \textbf{Training.} We train all models on 4 V100s with a batch size of 32. The learning rates of generator and discriminator are 0.0025 and 0.002, respectively. We set the gamma value of RGB image and semantic mask to 2 and 200. Such a high gamma for semantic masks prevents the semantic masks dominating the backward gradient which affects image quality. Following EG3D \cite{eg3d}, we regularize the generator pose conditioning by randomly swapping the conditioning pose of the generator with another random pose. We fit a pose distribution from the extracted poses by an off-the-shelf pose detector \cite{deng2019accurate}, and randomly sample a camera pose for swapping. We adopt the blur initialization and style mixing trick in StyleGAN3 \cite{karras2021alias}. 

We train our models in two stages split by the neural rendering resolution: 64 $\times$ 64 for 10M images and at 128 $\times$ 128 for additional 2.5M images. Note that we obtain the comparable performance with EG3D with only about half training iterations (25M + 2.5M for EG3D). 

\section{Experiment details.} 
\label{sec:2}

\subsection{Baselines}

We take a comparison with StyleNeRF \cite{gu2021stylenerf}, StyleSDF \cite{orel2021stylesdf}, FENeRF\cite{sun2021fenerf}, EG3D\cite{eg3d} and SofGAN\cite{chen2020sofgan}. For StyleNeRF and StyleSDF, we utilize their official codes and train on different resolutions or datasets for comparison. For FENeRF, we re-implement it following the paper and obtain a reasonable performance quantitatively and qualitatively. For EG3D,  we use the quantitative results in their paper for comparison. 

\section{Additional Visual results}
\label{sec:3}
\begin{figure*}[b]
  \centering
  \includegraphics[width=0.9\linewidth]{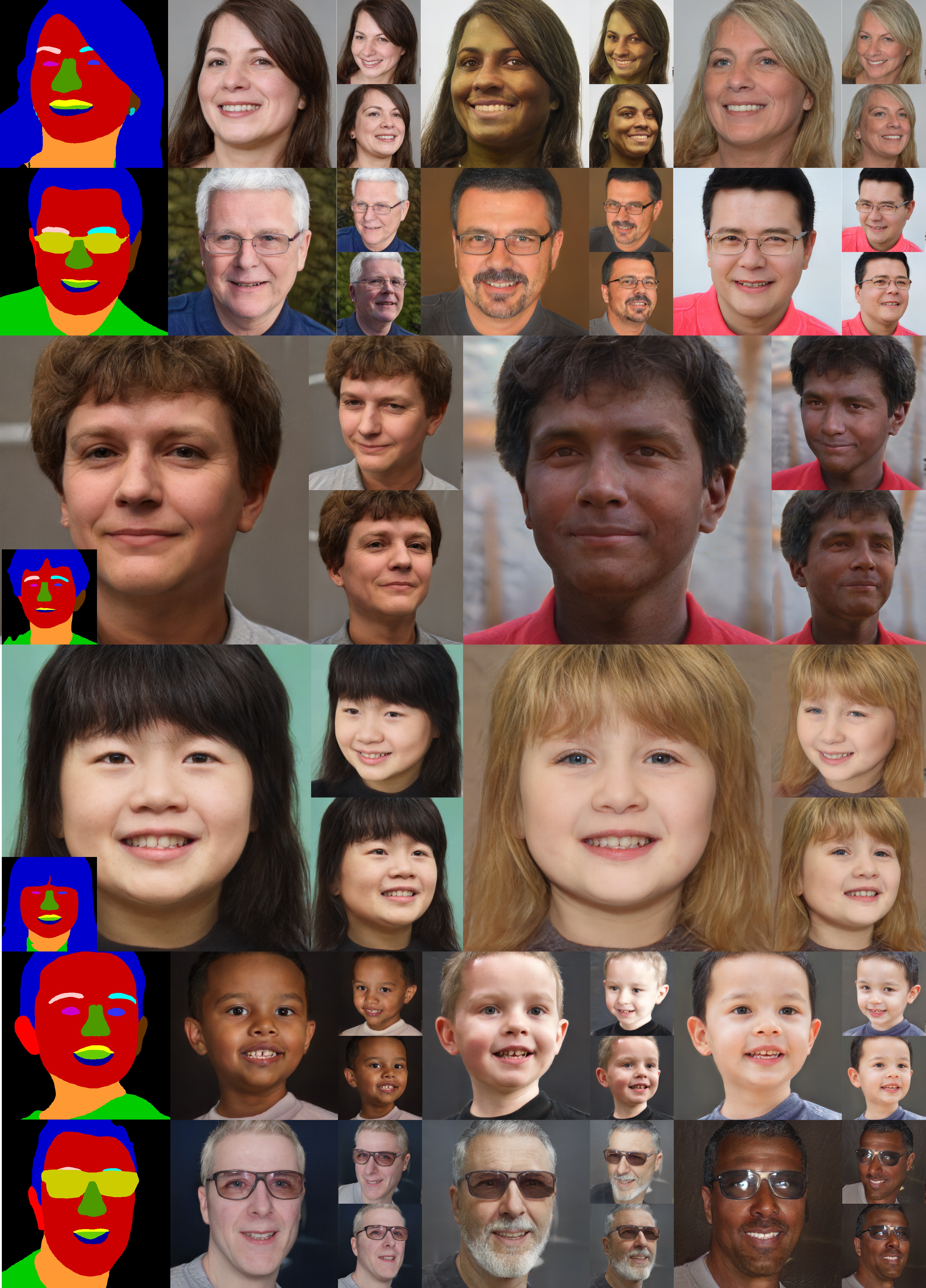}
  \caption{}
  \Description{}
  \label{fig:global_editing}
\end{figure*}

\begin{figure*}[b]
  \centering
  \includegraphics[width=0.9\linewidth]{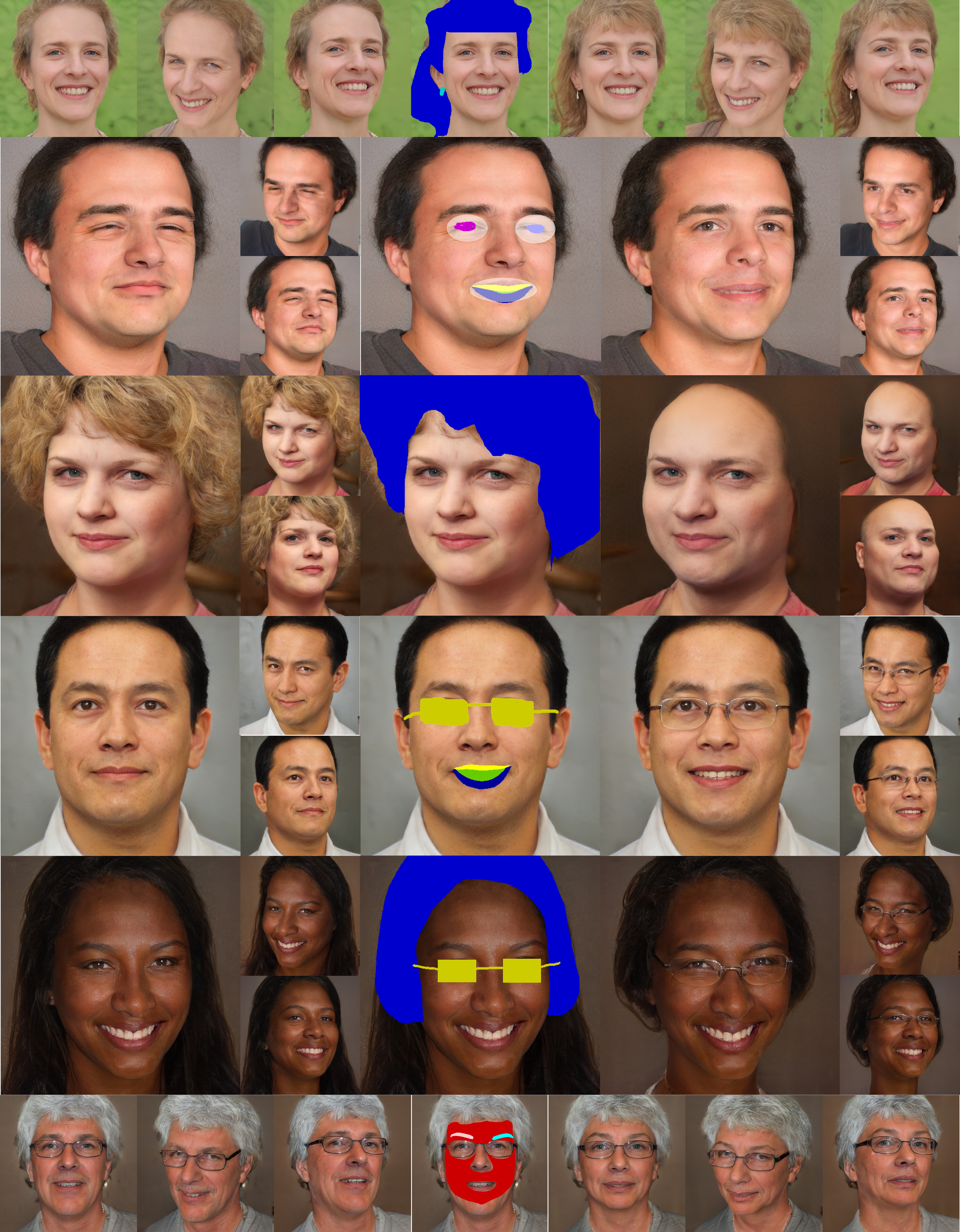}
  \caption{}
  \Description{}
  \label{fig:local_editing}
\end{figure*}

\begin{figure*}[b]
  \centering
  \includegraphics[width=0.9\linewidth]{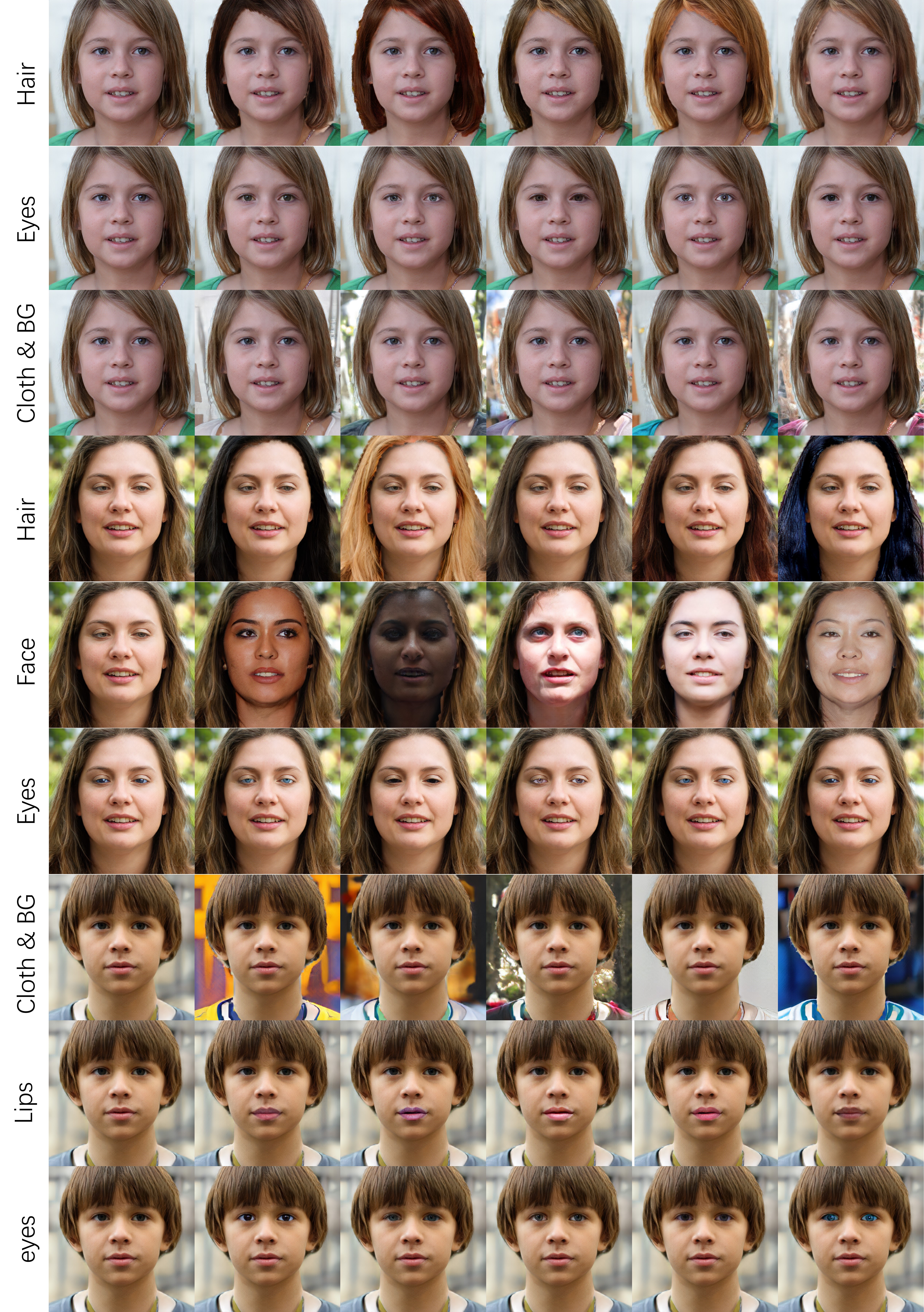}
  \caption{}
  \Description{}
  \label{fig:texture_editing}
\end{figure*}

\begin{figure*}[b]
  \centering
  \includegraphics[width=\linewidth]{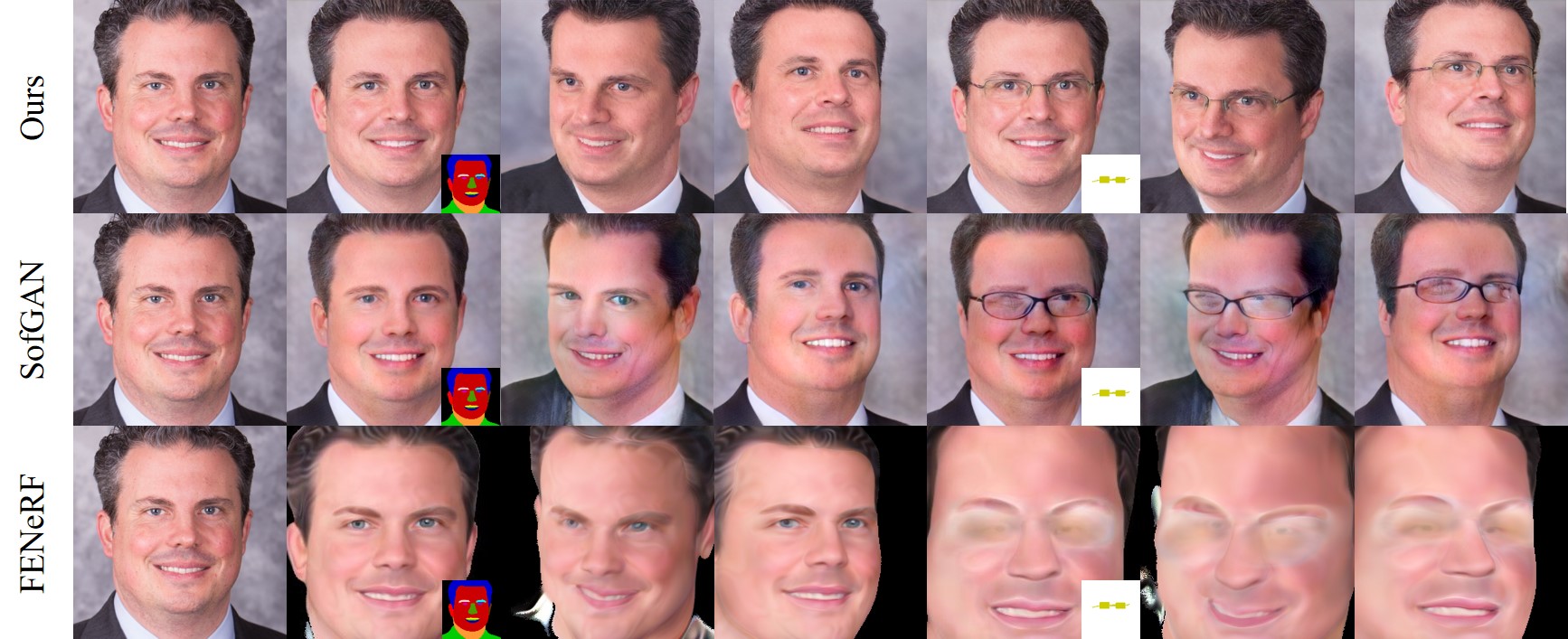}
  \caption{}
  \Description{}
  \label{fig:real_editing}
\end{figure*}

\begin{figure*}[b]
  \centering
  \includegraphics[width=\linewidth]{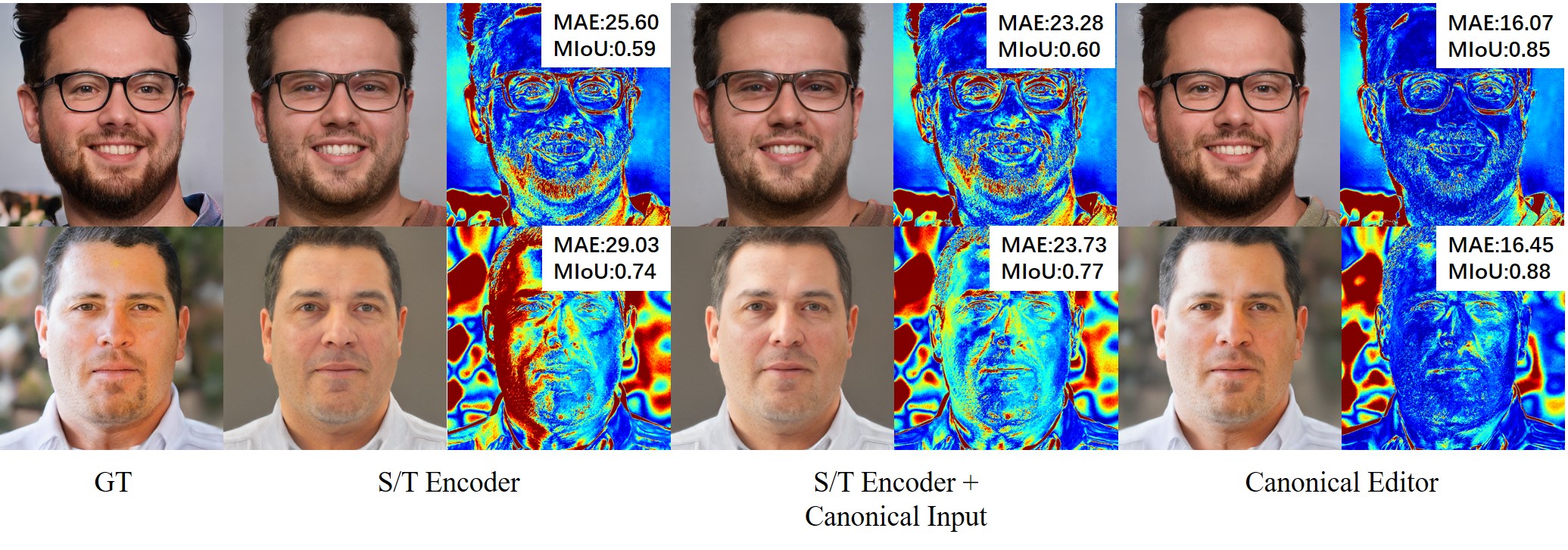}
  \caption{}
  \Description{}
  \label{fig:ablation_canonical}
\end{figure*}

\begin{figure*}[b]
  \centering
  \includegraphics[width=0.8\linewidth]{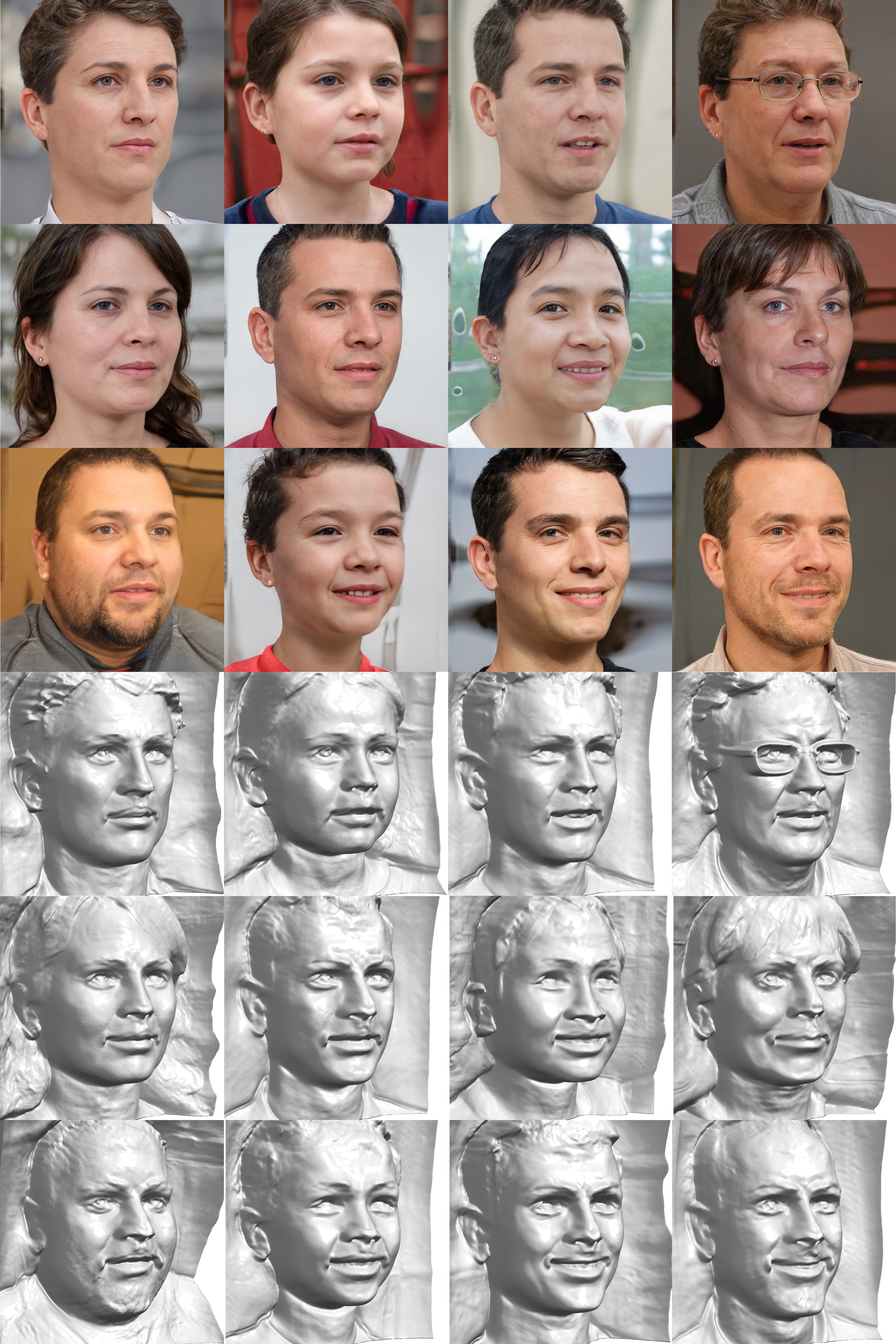}
  \caption{}
  \Description{}
  \label{fig:shape}
\end{figure*}

In this section we provide additional visual results of experiments related to the our core technical contributions: the global style adjustment (Fig.~\ref{fig:global_editing}), interactive local shape editing (Fig.~\ref{fig:local_editing}), region-level texture editing (Fig.~\ref{fig:texture_editing}), real image editing (Fig.~\ref{fig:real_editing}), the effectiveness of our proposed canonical encoder (Fig.~\ref{fig:ablation_canonical}) and the high-fidelity facial geometry (Fig.~\ref{fig:shape}).

\end{document}